\documentclass[journal]{IEEEtran}
\usepackage{xcolor,soul,framed} 

\colorlet{shadecolor}{yellow}
\usepackage[pdftex]{graphicx}
\graphicspath{{../pdf/}{../jpeg/}}
\DeclareGraphicsExtensions{.pdf,.jpeg,.png}

\usepackage[cmex10]{amsmath}
\usepackage{array}
\usepackage{mdwmath}
\usepackage{mdwtab}
\usepackage{eqparbox}
\usepackage{url}
\usepackage{multirow}
\usepackage{amsfonts}
\usepackage{bbding}
\usepackage{graphicx}
\usepackage{subfigure}
\usepackage{array}
\usepackage{amsmath,epsfig}
\usepackage{booktabs}
\usepackage{graphicx}
\usepackage[driverfallback=dvipdfm,
CJKbookmarks=true, bookmarksnumbered=true, bookmarksopen=true,
colorlinks=true,
pdfborder=001,
citecolor=blue, linkcolor=blue, anchorcolor=red, urlcolor=black]{hyperref}
\usepackage{float}
\begin{document}\sloppy

\title{MMNet: Multi-Collaboration and Multi-Supervision Network for Sequential Deepfake Detection}
    
\author{Ruiyang~Xia,~Decheng~Liu,~Jie~Li,~Lin~Yuan,~Nannan~Wang,~\IEEEmembership{Member, IEEE},~Xinbo~Gao,~\IEEEmembership{Senior Member, IEEE}
\IEEEcompsocitemizethanks{
\IEEEcompsocthanksitem

\emph{(Corresponding authors: Nannan Wang; Xinbo Gao.)}
\IEEEcompsocthanksitem R. Xia and J. Li are with the State Key Laboratory of Integrated Services Networks, School of Electronic Engineering, Xidian University, Xi'an 710071, Shaanxi, China
(e-mail: ryon@stu.xidian.edu.cn; leejie@mail.xidian.edu.cn).
\IEEEcompsocthanksitem D. Liu is with the State Key Laboratory of Integrated Services Networks, School of Cyber Engineering, Xidian University, Xi'an 710071, China (e-mail: dcliu.xidian@gmail.com).
\IEEEcompsocthanksitem L. Yuan is with the Chongqing Key Laboratory of Image Cognition, Chongqing University of Posts and Telecommunications, Chongqing 400065, China (e-mail:yuanlin@cqupt.edu.cn).
\IEEEcompsocthanksitem N. Wang is with the State Key Laboratory of Integrated Services Networks, School of Telecommunications Engineering, Xidian University, Xi'an 710071, Shaanxi, China
(e-mail: nnwang@xidian.edu.cn).
\IEEEcompsocthanksitem X. Gao is with with the Chongqing Key Laboratory of Image Cognition, Chongqing University of Posts and Telecommunications, Chongqing 400065, China
(e-mail: gaoxb@cqupt.edu.cn).
}
}

\markboth{}{Roberg \MakeLowercase{\textit{et al.}}: Bare Demo of IEEEtran.cls for Journals}

\maketitle

\begin{abstract}
Advanced manipulation techniques have provided criminals with opportunities to make social panic or gain illicit profits through the generation of deceptive media, such as forged face images. In response, various deepfake detection methods have been proposed to assess image authenticity. Sequential deepfake detection, which is an extension of deepfake detection, aims to identify forged facial regions with the correct sequence for recovery. Nonetheless, due to the different combinations of spatial and sequential manipulations, forged face images exhibit substantial discrepancies that severely impact detection performance. Additionally, the recovery of forged images requires knowledge of the manipulation model to implement inverse transformations, which is difficult to ascertain as relevant techniques are often concealed by attackers. To address these issues, we propose Multi-Collaboration and Multi-Supervision Network (MMNet) that handles various spatial scales and sequential permutations in forged face images and achieve recovery without requiring knowledge of the corresponding manipulation method. Furthermore, existing evaluation metrics only consider detection accuracy at a single inferring step, without accounting for the matching degree with ground-truth under continuous multiple steps. To overcome this limitation, we propose a novel evaluation metric called Complete Sequence Matching (CSM), which considers the detection accuracy at multiple inferring steps, reflecting the ability to detect integrally forged sequences. Extensive experiments on several typical datasets demonstrate that MMNet achieves state-of-the-art detection performance and independent recovery performance.
\end{abstract}


%
\IEEEpeerreviewmaketitle


\section{Introduction}

\IEEEPARstart{T}{he} development of deep learning not only brings significant improvement to traditional visual tasks \cite{yolov7,seg} but gives birth to massive novel and heuristic vision applications \cite{edit1,facelab,rennactment}. Deepfake, a new technique used to generate artificial media by deep neural networks \cite{5}, has raised public concerns about personal security and privacy. To fight against malicious facial deepfakes, detection methods have been naturally proposed and extensively studied over recent years. 
\begin{figure}
\includegraphics[width=\linewidth]{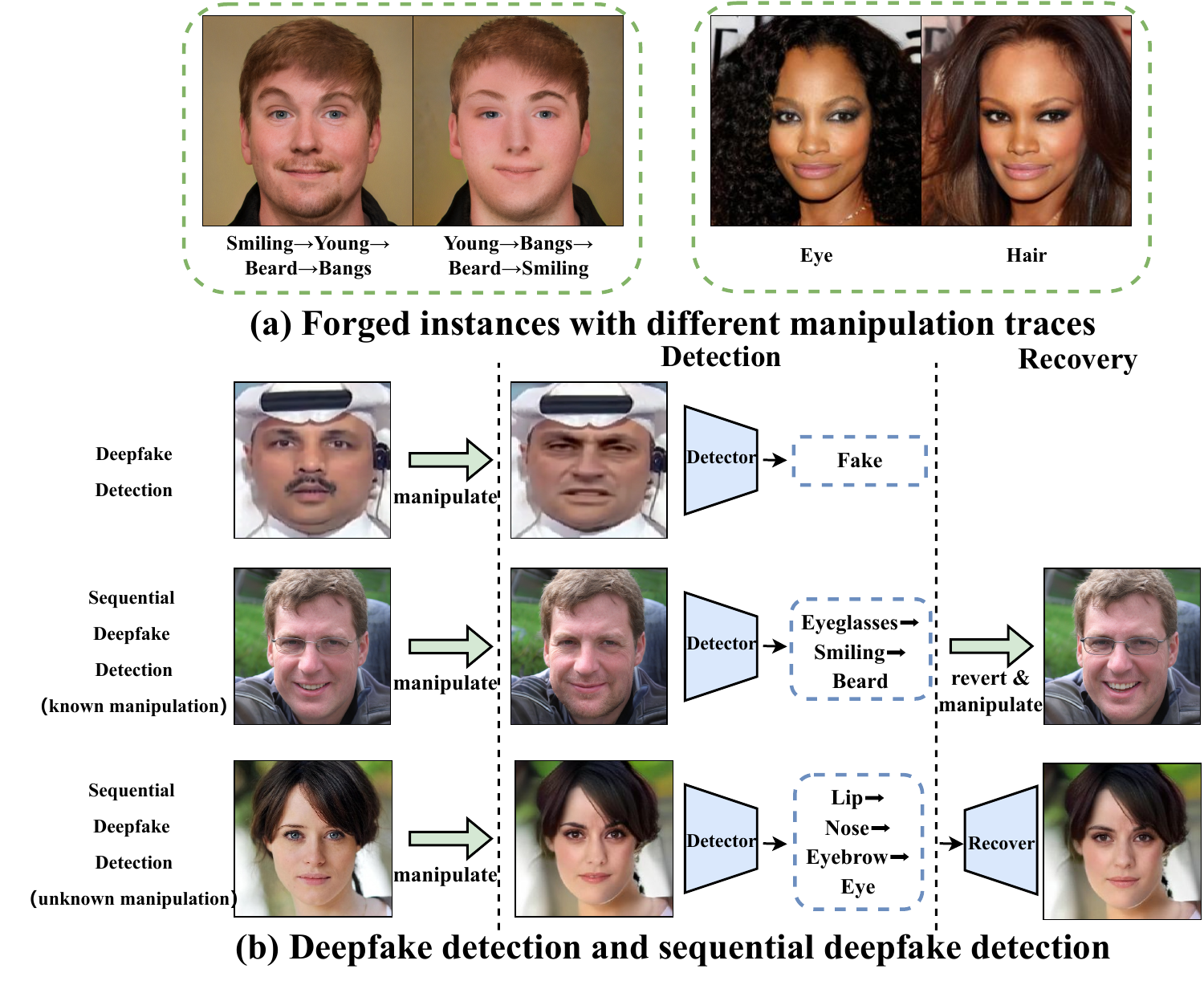}
\centering
\caption{(a) The comparison between the forged images with sequential manipulation trace (left) and spatial manipulation trace (right). (b) The difference between deepfake detection and sequential deepfake detection. If the manipulated model is known, reverting the sequence and manipulating again will achieve recovery. Independent recovery is adopted when the manipulation model is unknown.}
\label{f1}
\end{figure}
\par Conventional approaches to deepfake detection predominantly prioritize binary classification as a means of assessing the genuineness of facial imagery. However, these binary outcomes offer restricted insights for subsequent investigations, such as face recovery. Additionally, contemporary deepfake detection methods primarily concentrate on identifying face replacement and reenactment. Nevertheless, the editing techniques can also be used by the attacker to fabricate fictitious identities for the purpose of deceiving others, such as portraying a healthy leader as being unwell \cite{mit}.
\par The realm of face editing techniques encompasses a wide array of manipulated methods capable of distorting multiple facial regions to achieve deceptive visual outcomes. In addition, as shown in Fig. \ref{f1}(a), the entanglement of the generator in the latent vector space makes forged images different even if the manipulated region labels are the same \cite{stylemap, jiang2021talk}. Therefore, to identify these manipulated regions and achieve recovery for defense, sequential deepfake detection is proposed to detect these manipulated regions with the correct sequence \cite{6}. The dissimilarities between deepfake detection and sequential deepfake detection are summarized in Fig. \ref{f1}(b). 
\par A proficient sequential deepfake detector exhibits the capability to discern various combinations of spatial and sequential manipulation traces present within forged images. Previous advanced method deems sequential deepfake detection as a special image captioning task and adopts an enhanced auto-regressive model to detect manipulated regions \cite{6}. However, as only one network branch is available during detection, it is difficult to learn the representations of these various manipulation traces in the forged face images because the manipulation combinations have high discrepancies. In addition, to achieve recovery, an intuitive way is to use the same manipulation technique but the reverse manipulation sequence \cite{6}. Nonetheless, it is challenging to perceive which manipulation technique is utilized in the real scenario as the prevalence of numerous similar techniques and attackers may intentionally conceal their use of certain techniques. Lastly, previous evaluation matrices only focus on the detection accuracy in a single step rather than the whole continuous sequence. In spite of that, we prefer to know the matching degree between the complete predicted sequence and the corresponding sequential ground-truth so that aware of the quality of the generated face image in the recovering stage.
\par Albeit these forged facial images seem complex, a novel network called Multi-Collaboration and Multi-Supervision Network (MMNet) is proposed by introducing multiple detection branches and extra supervisions to generalize various manipulation traces. Moreover, MMNet can recover face images without the knowledge of manipulation techniques. To the best of our knowledge, we are the early exploration to simultaneously achieve detection and independent recovery, which can be seen as a baseline for further research. To reflect the capability of capturing integrally forged sequences and the potential recovering quality of the forged facial images, a new evaluation metric named Complete Sequence Matching (CSM) is proposed to evaluate the matching degree between the complete predicted sequence and the sequential ground-truth.

\par Therefore, the main contributions of this paper can be summarized as follow:
\begin{itemize}
    \item A novel network named MMNet is proposed for sequential deepfake detection. MMNet incorporates both multi-collaboration and multi-supervision modules, which work together to generalize across a range of manipulation traces and thus provide more accurate detection accuracy.
    \item Additionally, MMNet is the first network that attempts to recover manipulated facial regions even without prior knowledge of the specific manipulation technique.
    \item A new evaluation metric called Complete Sequence Matching (CSM) is proposed to expand the temporal dimension, treat the predicted results as an entity and then match with the ground-truth sequence. 
    \item Extensive experiments on several public typical manipulation datasets demonstrate the superior performance of our proposed method and the importance of the presented evaluation metric.
\end{itemize}

\section{Related Work}
\subsection{Deepfake Detection}
The aim of deepfake detection is to make a model able to mine the discrepancy of distribution between real and fake images. Therefore, the challenge of this task is to not only achieve high detection accuracy but also keep this characteristic among datasets with different manipulation techniques and image qualities. 
\par While numerous deepfake detection methods have been proposed in recent years, they can be broadly categorized into spatial-based, frequency-based, and data-based approaches. Spatial-based methods involve modifying the network architecture to enhance the extraction of spatial face image features \cite{recce, tifs2, tifs3}. Frequency-based methods focus on analyzing the frequency differences between real and fake images \cite{qian2020thinking,freq2,tifs1}. Data-based methods aim to identify common characteristics of fake images to improve model generalization across different manipulation techniques \cite{li2020face,shiohara2022detecting,pcl}.
\par The objective of both deepfake detection and sequential deepfake detection is to distinguish between authentic and counterfeit images \cite{6}. However, the latter involves a more intricate analysis of the correlation between detection and recovery processes. Specifically, sequential deepfake detection strives to pinpoint the manipulated facial regions and establish the optimal order of operations to recover the original identity-related features. Consequently, sequential deepfake detection represents a newly emerging and notably more demanding research task.
\subsection{Face Editing}
\par Facial editing encompasses the alteration of various attributes and components within an image that pertains to the face, including but not limited to the color of the hair or skin, gender, age, and the addition of glasses. 
\par In \cite{ICGAN}, the authors introduced a technique called the Invertible Conditional GAN (IC-GAN) to facilitate complex image editing. This method combines an encoder with a conditional GAN (cGAN) \cite{CGAN}. Another notable contribution by Lample et al. \cite{lample}, involves an encoder-decoder architecture that is specifically trained to disentangle salient information and attribute values in the latent space, thereby enabling image reconstruction. Building upon these advancements, \cite{stargan} introduced an enhanced approach known as StarGAN, which eliminates the need to develop separate models for each pair of image domains. Additionally, He et al. proposed attGAN in \cite{attgan}, which relaxes the attribute-independent constraint on the latent representation and instead focuses on applying the attribute-classification constraint to the generated image, ensuring accurate attribute changes. Moreover, based on the pretrained styleGAN \cite{ffhq}, more advanced techniques are proposed and produce more realistic edited images and some interesting applications \cite{stylemap,jiang2021talk}.
\begin{figure*}[t]
\begin{center}
\includegraphics[width=\linewidth]{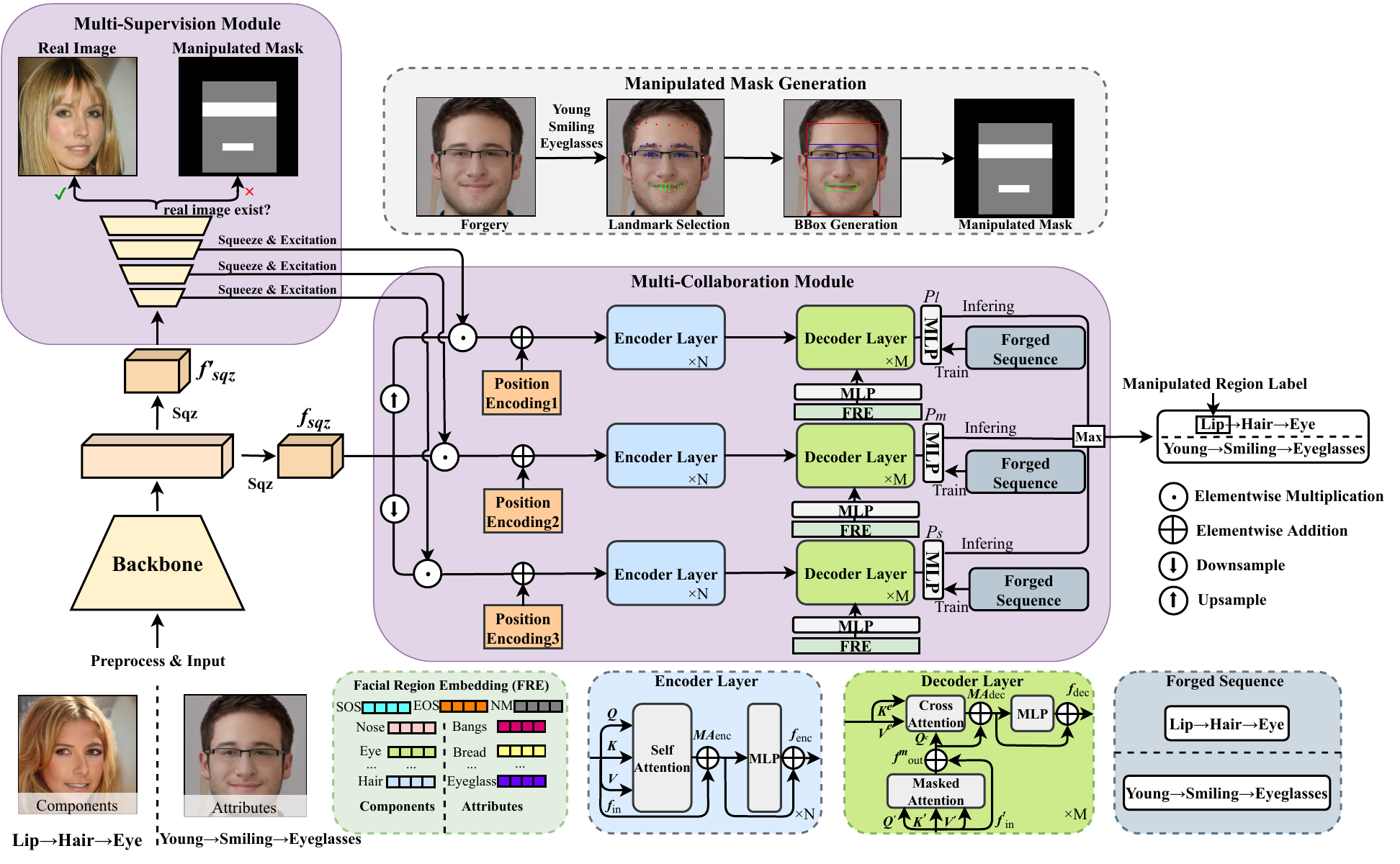}
\end{center}
\caption{The pipeline of the detection framework. The image features are extracted by a backbone. The squeezed features \textbf{$f_{sqz}$} are fed into the multi-collaboration module to generate different scale features. Facial region embeddings interact with different scale features in the decoders.  Different decoders collaborate to output manipulated region labels. The squeezed features \textbf{$f^{'}_{sqz}$} are processed in the multi-supervision module to bring external channel attention. SOS, EOS, and NM denote start-of-sentence, end-of-sentence, and no manipulation, respectively.}
\label{f2}
\end{figure*}
\par With the development of the editing technique, the authenticity of generated images has notably improved, accompanied by a heightened ease of utilizing these techniques. However, this progress inadvertently makes attackers manipulate forged images with greater ease, leading to the dissemination of misleading information and the humiliation of unsuspecting victims \cite{mit}.
\subsection{Image Captioning}
\par The goal of image captioning is to depict the global content and the local details of an image in natural language \cite{icsurvey}. The challenge of this task is both the design of the visual and language framework. To elevate the representation of the visual framework, different attention mechanisms are introduced such as general attention \cite{ga} and self-attention \cite{sa}. As for language framework, LSTM-based \cite{lstm,lstm1,lstm2} and CNN-based \cite{cnn} approaches are the main strategies to deal with this task at the early stage. In recent years, Transformer-based \cite{transformer,t1,t2,t3} and Image-Text early fusion methods \cite{bert,i2t,i2t2,i2t3} further achieve appealing captioning performance.
\par Given that forged face images may contain diverse manipulation traces, a visual encoder can be employed to extract spatial manipulation features, while a language model can focus on sequential manipulation traces. This makes sequential deepfake detection akin to a unique image captioning task, wherein the output is limited to keywords (i.e., manipulated region labels) obtained through scanning a forged face image. 
\section{Methodology}
\par In this section, we will introduce MMNet which consists of a detection framework and a recovery framework. The detection framework identifies manipulated facial components/attributes in the correct sequence, while the recovery framework is optional and used when the manipulation model is unknown. Additionally, we present a new evaluation metric called Complete Sequence Matching (CSM) that treats the predicted results as a complete entity and matches it with the sequential ground-truth.

\subsection{Detection Framework}
\par The detection framework is shown in Fig. \ref{f2}. It consists of a backbone for feature extraction, a multi-collaboration module to cope with spatial and sequential manipulation traces, and a multi-supervision module to promote the localization of manipulated regions. 
\par The challenge of learning representations for diverse spatial manipulation traces arises from various manipulated areas, which is evident in the context of a single-scale detection branch. Feature pyramid network (FPN) \cite{fpn} has been extensively employed for detecting objects of varying scales by leveraging inherent network features with different scales. However, the computational cost of FPN is exacerbated by the interactions among different network layers. Recent studies have highlighted that the effectiveness of FPN primarily stems from its multiple output structure \cite{yolof,vitdet}. This observation has motivated us to propose a concise version of FPN to address the issue of diverse spatial manipulation traces. Accordingly, we default to setting three detection branches to capture small (e.g., eye), middle (e.g., nose), and large (e.g., hair) manipulated regions. To reduce redundancy in the channel dimension, we initially squeeze the output features from the backbone. Subsequently, we respectively employ convolution and deconvolution operations with 2 strides to downsample and upsample the spatial size of the squeezed features $\boldsymbol{f_{\text {sqz}}}$. Hence, expanding the feature scale from 16 to 8, 16, and 32, respectively.
\par In each detection branch, to elevate the determination confidence, a self-attention mechanism is employed in the encoder layer for capturing long-range dependencies among different facial regions. Prior to training, independent position encodings are computed and added to their corresponding spatial features \cite{transformer}. The self-attentions are computed as follows:
\begin{equation}
\begin{gathered}
    \operatorname{Attn}\left(\boldsymbol{Q}_{\boldsymbol{h}}, \boldsymbol{K}_{\boldsymbol{h}}, \boldsymbol{V}_{\boldsymbol{h}}\right)=\operatorname{Softmax}\left(\frac{\boldsymbol{Q}_{\boldsymbol{h}} \boldsymbol{K}_{\boldsymbol{h}}{ }^T}{\sqrt{d}}\right) \boldsymbol{V}_{\boldsymbol{h}}, \\
    h=1, \ldots, H
\end{gathered}  
\end{equation}%
where $\boldsymbol{Q_h}$, $\boldsymbol{K_h}$ and $\boldsymbol{V_h}\in{\mathbb{R}}^{n^{2}\times d}$ denote the $h$-th head of query $\boldsymbol{Q}$, key $\boldsymbol{K}$ and value $\boldsymbol{V}$. In each attention head, $n$ and $d$ denote the spatial length and the number of channels. Multi-head attentions $\boldsymbol{MA_\text{enc}}\in{\mathbb{R}}^{n^{2}\times Hd}$ in the encoder layer are computed by concatenating these head attentions in the channel and adding the flattened extracted image features $\boldsymbol{f_\text{in}}\in \mathbb{R}^{n^2\times Hd}$:
\begin{equation}
\begin{aligned}
\boldsymbol{MA_{\text {enc}}} = & \operatorname{LN}(\text{ Concat }\left(\operatorname{Attn}\left(\boldsymbol{Q_1}, \boldsymbol{K_1}, \boldsymbol{V_1}\right), \ldots,\right. \\
& \left.\operatorname{Attn}\left(\boldsymbol{Q_H}, \boldsymbol{K_H}, \boldsymbol{V_H}\right)\right) \boldsymbol{W_e}^T + \boldsymbol{f_{\text {in}}}),
\end{aligned}
\end{equation}
where $H$ indicate the number of heads. LN denotes layer normalization. Encoder features $\boldsymbol{f_\text {enc}}\in{\mathbb{R}}^{n^{2}\times Hd}$ are hence expressed as:
\begin{equation}
\boldsymbol{f_\text {enc}}=\operatorname{LN\left(\operatorname{MLP_\text {enc}}\left(\boldsymbol{MA_\text {enc}}\right)+\boldsymbol{MA_\text {enc}}\right)},
\end{equation}
where $\operatorname{MLP_\text {enc}}$ indicates the multilayer perceptron in the encoder layer.
\par Accordingly, a single decoder may impede the sequential detection performance as different branches have distinct concentrations. As a result, there are also three decoder branches that correspond to the encoders. Within each decoder, learnable embeddings are utilized to represent labels for manipulated facial regions, no manipulation (NM), start-of-sentence (SOS), and end-of-sentence (EOS). The SOS and EOS indicate the beginning and end of prediction, while the NM is output when there is no manipulation at a certain step. The selected embeddings are subsequently input into the masked attention layer, which serves to model the sequential inference process:
\begin{equation}
\begin{gathered}
    \operatorname{MaskAttn}\left(\boldsymbol{Q^{'}_{\boldsymbol{h}}}, \boldsymbol{K^{'}_{\boldsymbol{h}}}, \boldsymbol{V^{'}_{\boldsymbol{h}}}\right)=\operatorname{Softmax}\left(\frac{\boldsymbol{Q^{'}_{\boldsymbol{h}}} \boldsymbol{K^{'}_{\boldsymbol{h}}{ }^T}}{\sqrt{d}}\right) \boldsymbol{M}\boldsymbol{V^{'}_{\boldsymbol{h}}}, \\
    h=1, \ldots, H
\end{gathered}  
\end{equation}%
where the $h$-th head query $\boldsymbol{Q^{'}_{\boldsymbol{h}}}$, key $\boldsymbol{K^{'}_{\boldsymbol{h}}}$ and value $\boldsymbol{V^{'}_{\boldsymbol{h}}}\in{\mathbb{R}}^{n_s \times d}$ come from the linear projection of the input embeddings $\boldsymbol{f^{'}_\text {in}}\in{\mathbb{R}}^{n_s\times Hd}$. $n_s$ is the number of steps and each step has a predetermined label during training. $\boldsymbol{M}\in{\mathbb{R}}^{n_s \times n_s}$ denotes the lower triangular matrix with the value set as one, which prevents the model from seeing the rear information at each step during training. Then the output $\boldsymbol{f^{m}_\text {out}}\in{\mathbb{R}}^{n_s\times Hd}$ is computed by concatenating these single head attentions, doing dot-product with the learned weight matrix $\boldsymbol{W_m}\in{\mathbb{R}}^{Hd \times Hd}$, and adding the $\boldsymbol{f^{'}_\text {in}}$:
\begin{equation}
\begin{aligned}
\boldsymbol{f^{m}_{\text {out}}} = & \operatorname{LN}(\text{ Concat }\left(\operatorname{MaskAttn}\left(\boldsymbol{Q}^{'}_1, \boldsymbol{K}^{'}_1, \boldsymbol{V}^{'}_1\right), \ldots,\right. \\
& \left.\operatorname{MaskAttn}\left(\boldsymbol{Q}^{'}_H, \boldsymbol{K}^{'}_H, \boldsymbol{V}^{'}_H\right)\right) \boldsymbol{W}_m^T + \boldsymbol{f}^{'}_{\text {in}}).
\end{aligned}
\end{equation}
\par After getting $\boldsymbol{f_{\text {enc}}}$, cross attention layer is utilized to interact the embeddings with the extracted image features, which can be expressed as:
\begin{equation}
\begin{gathered}
\operatorname{CrossAttn}\left(\boldsymbol{Q}_{\boldsymbol{h}}^c, \boldsymbol{K}_{\boldsymbol{h}}^e, \boldsymbol{V}_{\boldsymbol{h}}^e\right)=\operatorname{Softmax}\left(\frac{\boldsymbol{Q}_{\boldsymbol{h}}^c \boldsymbol{K}_{\boldsymbol{h}}^{e T}}{\sqrt{d}}\right) \boldsymbol{V}_{\boldsymbol{h}}^e, \\
h=1, \ldots, H
\end{gathered}
\end{equation}
where queries $\boldsymbol{Q}_{\boldsymbol{h}}^c\in \mathbb{R}^{n_s\times d}$ are the linear projection of $\boldsymbol{f^{m}_\text{out}}$. Keys $\boldsymbol{K_h^e}$ and values $\boldsymbol{V_h^e}$ are linearly projected from $\boldsymbol{f_\text {enc}}$, respectively. Each head cross-attention will also be concatenated and add $\boldsymbol{f^{m}_\text{out}}$ to get the multi-head decoder attentions $\boldsymbol{MA_\text {dec}}\in\mathbb{R}^{n_s\times Hd}$ by following:
\begin{equation}
\begin{aligned}
\boldsymbol{MA_{\text {dec}}} = & \operatorname{LN}(\text{ Concat }\left(\operatorname{CrossAttn}\left(\boldsymbol{Q}^{c}_1, \boldsymbol{K}^{e}_1, \boldsymbol{V}^{e}_1\right), \ldots,\right. \\
& \left.\operatorname{CrossAttn}\left(\boldsymbol{Q}^{c}_H, \boldsymbol{K}^{e}_H, \boldsymbol{V}^{e}_H\right)\right) \boldsymbol{W}_d^T + \boldsymbol{f}^{m}_{\text {out}}).
\end{aligned}
\end{equation}
\par After that, the decoder features $\boldsymbol{f_\text {dec}}\in\mathbb{R}^{n_s\times Hd}$ are expressed as:
\begin{equation}
\begin{aligned}
\boldsymbol{f_\text {dec}}=\operatorname{LN\left(\operatorname{MLP_\text {dec}}\left(\boldsymbol{MA_\text {dec}}\right)+\boldsymbol{MA_\text {dec}}\right)},
\end{aligned}
\end{equation}
where $\operatorname{MLP_\text {dec}}$ indicates the multilayer perceptron in the decoder. After inputting $\boldsymbol{f_{\text {dec}}}$ into the last MLP layer, the prediction $P$ in one of the decoder branches at $i$-th step is computed as:
\begin{equation}
\begin{gathered}
P(i, \boldsymbol{X})=\max \left(P\left(y_{i, 1} \mid y_1, \ldots, y_{i-1}, \boldsymbol{X}\right), \ldots,\right. \\
\left.P\left(y_{i, n_l} \mid y_1, \ldots, y_{i-1}, \boldsymbol{X}\right)\right),
\end{gathered}
\end{equation}
where $n_l$ denotes the number of predicted labels. $\boldsymbol{X}$ indicates the input extracted features.
\par Sequential manipulation traces derive from various manipulated permutations. It is imperative for different decoder branches to collaborate together as such allows for a holistic analysis of the sequential predictions. Specifically, the SOS embedding is input to different decoders to initiate the detection procedure. At each step, the manipulated region label with the highest confidence among the different scale decoder branches is selected and then acts the input of these decoders in the subsequent step. Therefore, the probability of the proposed detection model at the $n$-th step can be formed as:
\begin{equation}
P_d(n)=\prod_{i=1}^{n} \max \left(P_s(i, \boldsymbol{X_s}), P_m(i, \boldsymbol{X_m}), P_l(i, \boldsymbol{X_l})\right),
\end{equation}
where $\boldsymbol{X_s}$, $\boldsymbol{X_m}$ and $\boldsymbol{X_l}$ represent the extracted features from the small, middle, and large scale branches, respectively. $y$ is the selected manipulated region label. Furthermore, the utilization of this selection strategy facilitates the simultaneous operation of distinct decoder branches, thereby circumventing temporal increments throughout the collaborative procedure. Each prediction at each step is contingent upon the entirety of preceding prognosticated manipulated region labels. If we gather and scrutinize the assorted forecasted sequences generated by divergent decoder branches at the final phase, it would necessitate a computation time threefold lengthier than that demanded by the suggested strategy. This disparity arises from the dynamic variation in inputs to each decoder branch at every step, thus impeding the attainment of concurrent execution.
\begin{figure}[t]
\includegraphics[width=6cm]{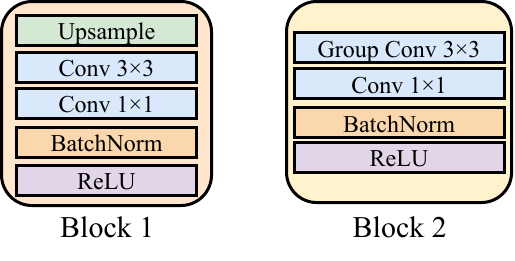}
\centering
\caption{The block structure of each spatial scale in multi-supervision module. The group number in Block 2 is equal to the number of input channels.}
\label{f8}
\end{figure}
\par Since the image-level supervisions (i.e., manipulated region labels) provide limited information, the learned representations may further benefit from the pixel-level supervisions as they show the difference in a specific image position. Therefore, a multi-supervision module is proposed by learning pixel-level supervision and offering external attention to the multi-collaboration module. A detailed illustration of each spatial scale block in the multi-collaboration module is presented in Fig. \ref{f8}. It consists of two blocks in each spatial scale, where the first block is responsible for increasing the spatial scale of features, while the second block further extracts the upsampled features. The squeezed features $\boldsymbol{f^{'}_{\text {sqz}}}$ go through these blocks and the intermediate scale features (8, 16, and 32 spatial sizes) are processed through the squeeze and excitation operation to get the external attention corresponding to the scales of detection branches, which can be expressed as:
\begin{equation}
\boldsymbol{Attn_{e}} = \operatorname{Sigmoid}\left(\boldsymbol{W}_{o}\operatorname{ReLU}\left(\boldsymbol{W}_{i} \boldsymbol{z}\right)\right),
\end{equation}
where $\boldsymbol{W}_{i} \in{\mathbb{R}}^{\frac{c_{i}}{r} \times c_{i}}$ and $\boldsymbol{W}_{o} \in{\mathbb{R}}^{c_{o} \times \frac{c_{i}}{r}}$. $c_{i}$ and $c_{o}$ denotes the feature channels in the multi-supervision module and the multi-collaboration module, respectively. $z$ indicates the specific scale features extracted from the multi-supervision module after global average pooling in the spatial dimension. $r$ follows the default configuration setting \cite{hu2018squeeze}.
\par Two pixel-level supervisions are utilized during training. The first is the real face images. The aim of the multi-supervision module is to learn the projection from the manipulated images to the real counterparts. Therefore, external attention not only brings the explicitly manipulated position between the real and fake image but also the sequential difference information for forged images with the same manipulated regions. The second is the manipulated mask related to the input images. To generate the mask, in Fig. \ref{f2}, the facial landmarks related to the manipulated region labels are selected to generate the bounding boxes. The values within these boxes vary, as larger manipulated regions have lower manipulated density and vice versa. Although the manipulated masks only provide relatively coarse position information compared to real face images, they can be adopted in a more severe situation that lacks the real images with respect to the fake one during training. Therefore, the real face images are the default selection unless they are not available. 

\par The detection framework is trained with the sum of cross-entropy loss from three detection branches ($\mathcal{L}_{\text{CE}}^s$, $\mathcal{L}_{\text{CE}}^m$, and $\mathcal{L}_{\text{CE}}^l$) and mean square error (MSE) $\mathcal{L}_{\text{MSE}}$ between the generated image/mask and the corresponding pixel-level supervision, which is expressed as:
\begin{equation}
\mathcal{L_{\text {Detection}}}=\mathcal{L}_{\text{CE}}^s+\mathcal{L}_{\text{CE}}^m+\mathcal{L}_{\text{CE}}^l+\mathcal{L}_{\text {MSE}}.
\end{equation}%
\begin{figure}
\begin{center}
\includegraphics[width=\linewidth]{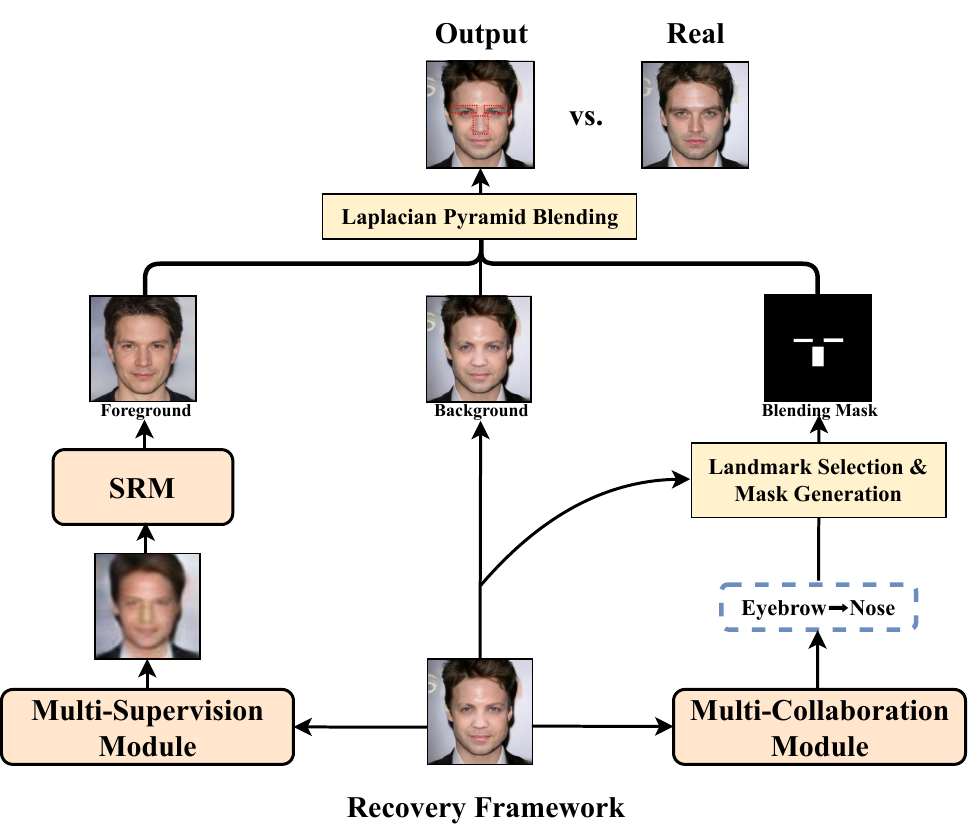}
\end{center}
\caption{The pipeline of recovery framework. MMNet receives a forged face image to output the manipulated region labels and the generated image, respectively. The forged image and the clearly generated image from SRM acts as background and foreground, respectively. After getting the blending mask, recovering face image is generated by the laplacian pyramid blending algorithm. The red box regions indicate the recovery result.}
\label{f3}
\end{figure}

\subsection{Recovery Framework}
\par The recovery framework serves to independently recover the forged face images. It applies to the scenario that the corresponding manipulation model is concealed by attackers. 
\par
\begin{figure}[t]
\includegraphics[width=5cm]{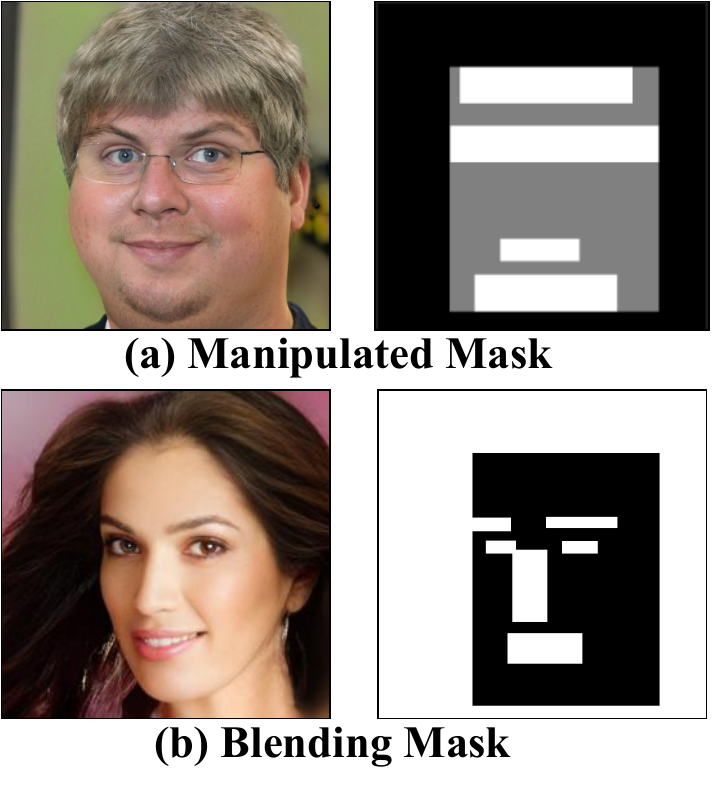}
\centering
\caption{Difference between two different generated masks. (a) Manipulated mask related to all region labels. (b) Blending mask related to all region labels.}
\label{f9}
\end{figure}

 The pipeline of the recovery framework is depicted in Fig. \ref{f3}. Due to the presence of various manipulation traces in forged face images, achieving direct projection from forgery to real using only the multi-supervision module is challenging. Therefore, images generated from the multi-supervision module will first be clarified as foreground by adopting a super-resolution module (SRM). The forged image then serves as the background because the editing technique mainly focuses on the designated regions, and the rest regions are still close to the real. To blend these two images, we collect the predictions from the detection framework and the facial landmarks related to the manipulated region labels from the forged face image. A blending mask is formed by finding the bounding boxes related to the selected landmarks. The generation of the blending mask is similar to the manipulated mask except for the value within the selected region is one and the index of the selected landmarks corresponds to the manipulated region labels. Moreover, for the hair region, we select the outside region of the selected bounding box. The difference between manipulated mask and blending mask related to all region labels is shown in Fig. \ref{f9}. 
\par After obtaining the manipulated image, the clearly generated image, and the blending mask, we adopt the laplacian pyramid blending algorithm to get the blending details and edges by following:
\begin{equation}
L^k(i, j)=\text{G}\left(M^k(i, j)\right)L_f^k(i, j)+\left(1-\text{G}\left(M^k(i, j)\right)\right)L_b^k(i, j),
\end{equation}
where $(i, j)$ denotes a specific image position. In the $k$-th scale, $M^k$ is the blending mask. G represents the gaussian smoothing function. $L_f^k$ and $L_b^k$ stand for the laplacian image of the foreground and background, which are computed by following: 
\begin{equation}
L_f^k(i, j)=F^k(i, j)-\text{Upsample}\left(\text{G}(F^{k-1}(i, j))\right),
\end{equation}
\begin{equation}
L_b^k(i, j)=B^k(i, j)-\text{Upsample}\left(\text{G}(B^{k-1}(i, j))\right),
\end{equation}
where $F^k$ and $B^k$ denote the foreground (i.e., clearly generated image) and background (i.e., forged image) in $k$-th spatial scale, respectively. Therefore, the $k$-th scale blending image $BI^k$ can be expressed as:
\begin{equation}
BI^k(i, j)=L^k(i, j)+\text{Upsample}\left(BI^{k-1}(i, j)\right).
\end{equation}
Note that the smallest scale of the blending image $BI^0$ is directly generated by using $M^0$ on the foreground and background. It is worthy note that if the face image is detected as real by the detection framework, it will not go through the recovery framework. 
\par We choose PixelStylePixel \cite{p2p}, a popular image translation model, as the SRM in the recovery framework. In the training stage, we combine the downsampled real images (from 2-16$\times$ downsample rate) and the generated images from the multi-supervised image as the training set. The recovery loss function is defined as:
\begin{equation}
\mathcal{L_{\text {Recovery}}}=\mathcal{L_{\text {SRM}}}=\lambda_1 \mathcal{L}_\text {MSE}+\lambda_2 \mathcal{L}_{\text {LPIPS}}+\lambda_3 \mathcal{L}_{\mathrm{ID}}+\lambda_4 \mathcal{L}_{\mathrm{reg}},
\end{equation}
where $\lambda_{1}$ to $\lambda_{4}$ are the item weights. The loss items respectively measure the difference of the pixel, feature perception, and face identity between the output of SRM and real images. $\mathcal{L}_{\mathrm{reg}}$ aims to reduce the distance between the latent vector to the average one.
\begin{figure}[t]
\begin{center}
\includegraphics[width=\linewidth]{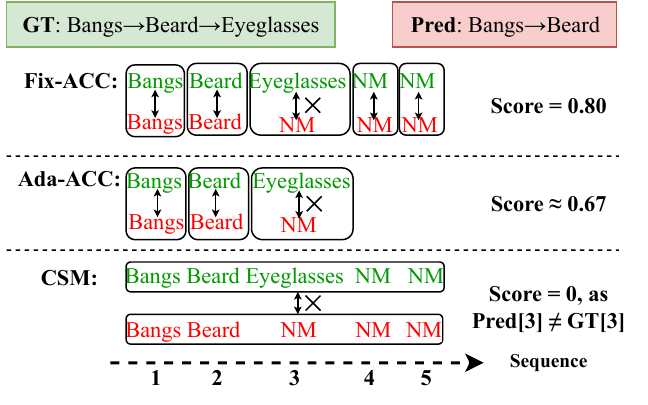}
\end{center}
\caption{Comparison among three evaluation metrics (a) Fixed Accuracy (b) Adaptive Accuracy, and (c) Complete Sequence Matching. NM denotes no manipulation.}
\label{f4}
\end{figure}
\subsection{Evaluation Metrics}
\par The comparison between existing evaluation metrics and CSM is shown in Fig. \ref{f4}. It can be seen that the fixed and adaptive accuracy all only reflect the detection accuracy per time. There is a vacancy to evaluate the model detection performance under multiple continuous steps. Therefore, CSM is proposed to represent the matching degree between the complete predicted sequence and the sequential ground-truth. In addition, CSM not only reflects the detection accuracy in continuous steps but also the potential recovering ability for forged face images. This is because of the demonstration that the best recovering performance appears when a predicted sequence is totally equal to the sequential ground-truth \cite{6}. Assuming the recovering quality only has two categories, i.e., completely consistent and inconsistent, the result of CSM will directly reflect the recovering ability without using the corresponding facial editing techniques. Therefore, the proposed metric for a single sample can be expressed as:
\begin{equation}
\text{CSM}( pred, g t)= \begin{cases}1, & pred[i]=g t[i], \forall i \in[1,2, \ldots, T] \\ 0, & \text { otherwise }\end{cases}
\end{equation}
where $T$ is the max length of the forged sequence.

\begin{table*}
    \centering
    \caption{Fixed accuracy (\%), adaptive accuracy (\%), and CSM (\%) comparisons of MMNet and other state-of-the-art detection methods on facial components and attributes manipulation datasets. In each column, the highest score is remarked in bold.}
    \setlength{\tabcolsep}{0.5cm}{
    \begin{tabular}{c|ccc|ccc}
        \hline
        \multirow{2}{*}{Method}  & \multicolumn{3}{c|}{Facial Components} & \multicolumn{3}{c}{Facial Attributes}\\
        \cline{2-7}
         & Fix-ACC & Ada-ACC & CSM & Fix-ACC & Ada-ACC & CSM\\
         \hline
         \hline
        DRN \cite{drn}&	66.06&	45.79&	-& 64.42&	43.20&	-\\
        DETR \cite{detr}& 69.75& 49.84&	- &67.62& 47.99 & -\\
        MA \cite{zhao2021multi}&	71.31& 52.94&	-&	67.58&	47.48&	-\\
        Two-Stream \cite{twostream}&	71.92&	53.89&	-& 66.77&	46.38&	-\\
        SeqFakeFormer w/o SECA  \cite{6}&	72.18&	54.64&	- &68.17&	48.81&	-\\
        SeqFakeFormer w SECA\cite{6}&	72.65&	55.30&	48.30 & 68.86&	49.63&	51.80\\
        \hline
        MMNet&	\textbf{73.93}&	\textbf{56.83}&\textbf{50.19} &\textbf{69.27}&	\textbf{50.44}&	\textbf{52.50}	\\
        \hline
    \end{tabular}}
    \label{1}
\end{table*}

\begin{table*}
    \centering
    \caption{Ablation study over different combinations of the multi-collaboration module on several manipulation validation sets.}
    \setlength{\tabcolsep}{0.6cm}{
    \begin{tabular}{c|ccc|ccc}
        \hline
        \multirow{2}{*}{Combination}  & \multicolumn{3}{c|}{Facial Components} & \multicolumn{3}{c}{Facial Attributes}\\
        \cline{2-7}
         & Fix-ACC & Ada-ACC & CSM & Fix-ACC & Ada-ACC & CSM\\
         \hline
         \hline
        Baseline &	69.87	&51.56	&45.12&67.02	&47.65	&50.12\\
        ME+SD 	&	70.30	&52.21	&45.78&67.56	&48.15	&50.88\\
        ME+MD (Mean) 	&70.68	&52.52	&46.23&67.85	&48.40	&51.17 \\
        ME+MD (Max) 	&\textbf{71.26}&\textbf{52.78}&\textbf{46.88}&\textbf{68.16}&\textbf{48.78}	&\textbf{51.58}\\
        \hline
    \end{tabular}}
    \label{2}
\end{table*}

\section{Experiments}
\subsection{Experimental Setup}
\subsubsection{Datasets.} \par Experiments are conducted on public sequential deepfake datasets: facial components and attributes manipulation datasets, which include 35,166 and 49,920 images, respectively. Both datasets were split into training, validation, and testing sets in an 8:1:1 ratio. The facial components dataset has $\{$Lip, Eye, Eyebrow, Hair, Nose$\}$ manipulated region labels with 28 sequential permutations, while the facial attributes dataset has $\{$Bangs, Eyeglasses, Beard, Smiling, Young$\}$ manipulated region labels with 26 permutations. Each region is only manipulated once and hence the length of manipulation sequence is range from 0 to 5. The real images \cite{ffhq, celeb} are manipulated by utilizing advanced editing techniques \cite{stylemap, jiang2021talk}. More details are described in \cite{6}.
\subsubsection{Implementation details for the detection framework}  All face images are resized into 256, and augmented by random JPEG compression (60-100 quality factor) only. The backbone is ResNet50 \cite{resnet} and the channel dimension is 256 for encoder and decoder layers in each branch. We utilize 4 attention heads in both the encoder and decoder and set the encoder and decoder layers to 2, which is the same as SeqFakeFormer. In the training stage, the optimizer is AdamW with $10^{-4}$ weight decay. The learning rate is set as $10^{-3}$ and divided by 10 after 70 and 120 epochs. The batch size is 64 and the number of epochs is set as 150. In the multi-supervision module, as the facial attributes dataset does not offer real images, we hence adopt the real images and the manipulated masks on the facial components and attributes dataset, respectively. As to the manipulated mask, for simplicity, there are only three different value settings (i.e., 0, 0.5, 1) for unmodified region, large (Young), and small (Bangs, Eyeglasses, Beard, and Smiling) manipulated regions. Facial landmarks are detected by using the Dlib library and the indexes of the selected landmarks corresponding to the five manipulated region labels, namely \{Bangs, Eyeglasses, Beard, Smiling, Young\}, are \{[69-74, 76, 77, 80, 81], [1, 16, 18-28, 37-48], [6-12], [49-68], [1-17, 69-81]\}.
\subsubsection{Implementation details for the recovery framework} The number of channel dimensions in the multi-supervision module for different scale features is 256. The SRM is pertrained on FFHQ \cite{ffhq} dataset and finetune on the generated face images from the proposed multi-supervision module and the downsampled real face images from \cite{celeb}. The training strategy and hyperparameter settings in $\mathcal{L_{\text {Recovery}}}$ are the same as \cite{5} during training. The batchsize is 8 and the number of iterations is set as $5 \times 10^{5}$. Since only the facial components dataset provides corresponding real images, the proposed recovery framework focuses exclusively on the dataset. As for the blending mask, the indexes of the selected landmarks corresponding to the manipulated region labels (i.e., \{nose, eye, eyebrow, lip, hair\}) are $\{$[28-36], [[37-42] (left eye), [43-48] (right eye)], [[18-22] (left eyebrow), [23-27] (right eyebrow)], [49-68], [1-17, 69-81]$\}$. The number of scales in the laplacian pyramid blending is 8, which is range from 2 to 256 with exponential multiples of 2. All models are implemented with Pytorch on NVIDIA GeForce RTX 3090. 

\begin{table*}
    \centering
    \caption{Ablation study over different combinations between the multi-supervision module and multi-collaboration module on facial components and attributes validation sets.}
    \setlength{\tabcolsep}{0.4cm}{
    \begin{tabular}{cc|ccc|ccc}
        \hline
        \multicolumn{2}{c|}{Combination} & \multicolumn{3}{c|}{Facial Components} & \multicolumn{3}{c}{Facial Attributes}\\
        \cline{1-8}
         Multi-Supervision & Multi-Collaboration& Fix-ACC & Ada-ACC & CSM & Fix-ACC & Ada-ACC & CSM\\
         \hline
         \hline
        -&-&	69.87	&51.56	&45.12&67.02	&47.65	&50.12 \\
        $\checkmark$&-&	70.82&	52.54&46.36&	67.88&	48.43&	50.97\\
        -&$\checkmark$& 71.26 &52.78 &46.88 &68.16 &48.78	&51.58 \\
        $\checkmark$&$\checkmark$&	\textbf{71.70}&	\textbf{53.04}&\textbf{47.19}&	\textbf{68.62}&	\textbf{49.30} &	\textbf{51.90} \\
        \hline
    \end{tabular}}
    \label{3}
\end{table*}

\subsection{Evaluation on the detection performance}
\subsubsection{Comparison with state-of-the-art} To establish the superiority and efficacy of our proposed method, we conducted comprehensive comparisons with state-of-the-art detectors that are widely recognized as cutting-edge solutions in the domain of sequential deepfake detection. 
\par Table \ref{1} showcases the evaluation metrics of fixed accuracy (Fix-ACC), adaptive accuracy (Ada-ACC), and CSM, along with the corresponding results obtained from the comparative analysis. These rigorous evaluations provide a robust validation of the exceptional performance of our proposed method. DRN \cite{drn}, MA \cite{zhao2021multi}, DETR \cite{detr}, and Two-Stream \cite{twostream} models consider each manipulation sequence as a separate class and treat sequential deepfake detection as a multi-class classification task. However, these approaches neglect the crucial analysis of forged sequential information, which makes the detection performance lower than sequential inferring models. Furthermore, as the number of different sequential permutations increases, these models would require an expanding number of classification branches, resulting in increased parameters. In contrast, SeqFakeFormer and MMNet can achieve better detection performance, which reflects the importance of addressing both spatial and sequential forged information.
\begin{figure}
\begin{center}
\includegraphics[width=\linewidth]{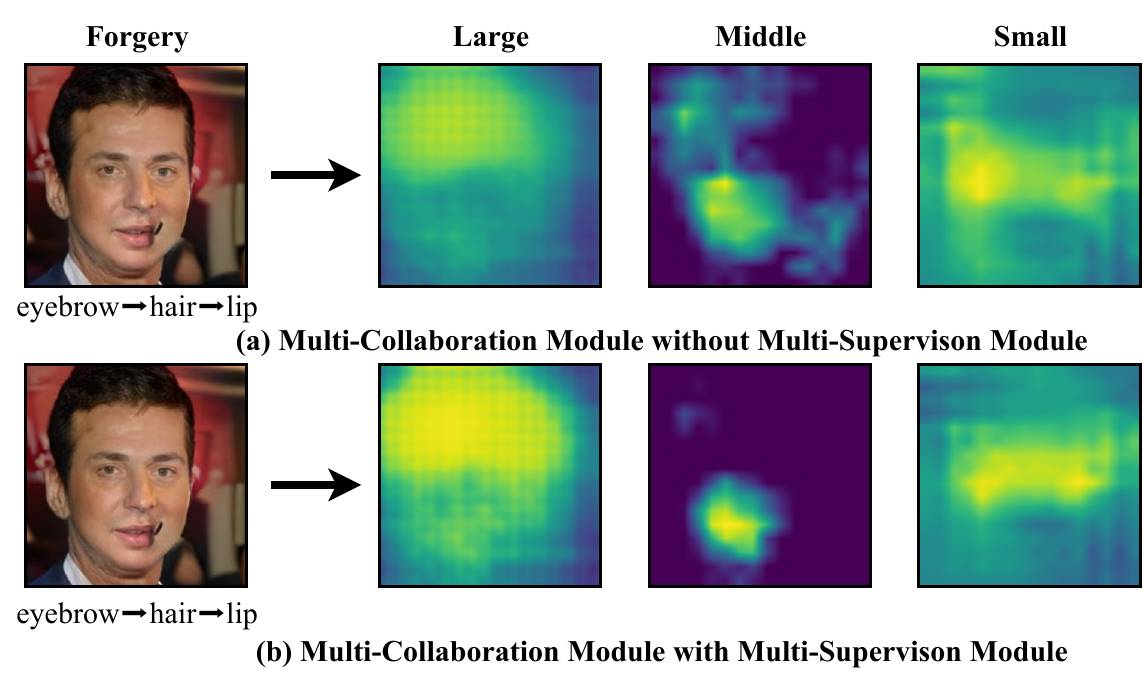}
\end{center}
\caption{Visualization of feature responses in different scale encoder branches without/with Multi-Supervison Module. Large, Middle, and Small indicate the large scale, middle scale, and small scale detection branches, respectively.}
\label{f5}
\end{figure}
\par Particularly, MMNet surpasses SeqFakeFormer in terms of Fix-ACC (1.28\% and 0.41\%), Ada-ACC (1.53\% and 0.81\%), and CSM (1.89\% and 0.70\%) on facial components and attributes datasets. These findings highlight the superior performance of the proposed MMNet, which integrates multi-collaboration and multi-supervision modules, in effectively capturing manipulated clues and achieving exceptional results in sequential deepfake detection.

\subsubsection{Discussion on the proposed modules} MMNet is mainly composed of multi-collaboration and multi-supervision modules. We first analyze the effectiveness of the multi-collaboration module. Table \ref{2} shows the detection results with different combinations. ME+SD and ME+MD denote the multiple encoders with a single decoder and multiple encoders with multiple decoders, respectively. The baseline is the single encoder with a single decoder. The gain of ME+SD indicates the structure of multiple encoders can better generalize different scales of spatial manipulation traces as each branch only responds to several specific discriminative manipulated features. Fig. \ref{f5} illustrates the feature responses from different scale encoder branches. The large scale detection encoder shows high responses in the hair region, the middle scale encoder focuses on the lip region, and the small scale encoder concentrates on the eyebrow region around. 
\par With multiple decoders, various sequential manipulation traces can be further captured and hence elevate the sequential detection performance. Notably, we have outlined two collaboration strategies in Table \ref{2}. The first strategy, termed "Mean", involves computing the average value of the predictions from different decoder branches, and then selecting the manipulated region label with the highest score at each step. The second strategy, termed "Max", follows Eq. (9) and Eq. (10). Our results reveal that adopting the max selection strategy yields better prediction results. We attribute this difference to the reason that the mean selection strategy may introduce noise as each step has one manipulation operation at most and each decoder exists a specific concentration, resulting in a decrease in the score of the correct label during the average process.
\par To assess the effectiveness of the proposed multi-supervision module, as shown in Table \ref{3}, ablation studies were conducted with different combinations of the proposed modules. It is evident from the results that the incorporation of the multi-supervision module leads to further improvement in the detection performance. This can be attributed to the promotion of forgery localization, which introduces external channel attention to the multi-collaboration module. To visually illustrate this, Fig. \ref{f5} also presents a comparison of feature responses between the multi-collaboration module without/with the multi-supervision module. The introduction of the multi-supervision module has resulted in a notable improvement in the concentration of feature responses within locally manipulated regions and hence contributes to a higher level of detection accuracy. 
\begin{figure}[t]
\includegraphics[width=\linewidth]{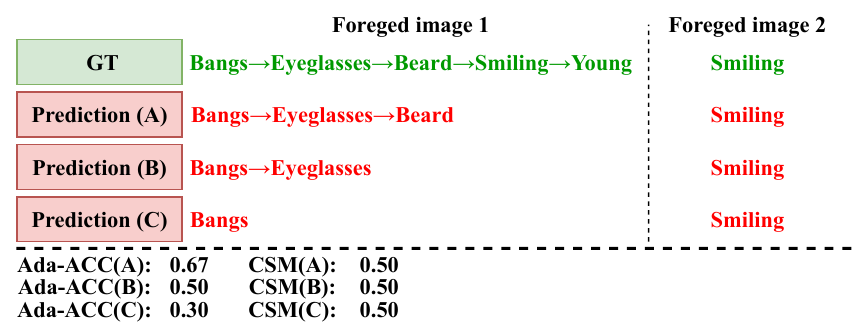}
\caption{Statistical example in facial attributes dataset. A, B, and C indicate different detection models.}
\label{f10}
\end{figure}
\subsubsection{Discussion on the proposed evaluation metric} Upon analyzing the results obtained from Ada-ACC and CSM, further insights into the properties of the model have been revealed. Specifically, the value of Ada-ACC is larger than the CSM score on the facial components dataset suggesting that the model has limitations in effectively capturing continuously manipulated sequences. Conversely, a higher CSM value in the facial attributes dataset suggests that the model may tend to miss certain manipulated regions, as the CSM value remains unchanged while the Ada-ACC result decreases. Our findings are supported by Table \ref{7}, which indicates that the detector is more likely to overlook manipulated regions, as evidenced by the number of samples in which the predicted sequence length is lower than the sequential ground-truth. Additionally, Fig. \ref{f10} provides a statistical example of instances where the detector ignores manipulated regions, leading to lower values of Ada-ACC as compared to CSM. These findings underscore the importance of evaluating both Ada-ACC and CSM as they can guide further research toward improving the model detection performance, which further highlights the necessity of the proposed evaluation metric CSM in our study.

\begin{table}[ht]
    \centering
    \caption{Statistical samples in which the length of the predicted sequence is larger and lower than the length of sequential ground-truth in the facial attributes dataset.}
    \setlength{\tabcolsep}{0.55cm}{
    \begin{tabular}{c|c|c}
        \hline
        Method & len(Pred)$\textgreater$len(GT)& len(Pred)$\textless$len(GT)\\
         \hline
         \hline
        MMNet &	572&	598\\
        \hline
    \end{tabular}}
    \label{7}
\end{table}

\subsubsection{Discussion on the pixel-level supervision} In order to evaluate the impact of various pixel-level supervisions on detection performance, on the facial components dataset, we adopt the default manipulated mask settings wherein unmodified regions are set to 0, large manipulated regions (i.e., hair) are set to 0.5, and small manipulated regions (i.e., nose, eye, eyebrow, and lip) are set to 1. The results of our comparative analysis of pixel-level supervision for both real images and manipulated masks are presented in Table \ref{9}. It can be observed that there is a difference of approximately 1\% in performance for Ada-ACC and CSM when manipulated masks and real images are used as pixel-level supervision, respectively. This difference in performance provides quantitative evidence for the superiority of using real images as the default pixel-level supervision when they are available.
\begin{table}[h]
    \centering
    \caption{Ablation study on different pixel-level supervisions on facial components test set.}
    \setlength{\tabcolsep}{0.75cm}
    \begin{tabular}{c|cc}
        \hline
        \multirow{2}{*}{pixel-level supervision}  & \multicolumn{2}{c}{Facial Components}\\
        \cline{2-3}
          & Ada-ACC & CSM \\
         \hline
         \hline
        manipulated mask & 55.79 & 49.12	\\
        real image &	\textbf{56.83}&	\textbf{50.19}\\
        \hline
    \end{tabular}
    \label{9}
\end{table}
\subsubsection{Discussion on the trade off between the spatial size and channel dimension} Due to the limited GPU memory, the multi-supervision module produces an output of 64 instead of 256 spatial size. To compensate for this, we increase the spatial size of the output at the expense of squeezing the channel dimension. Our analysis of Table \ref{8} reveals that despite the increase in spatial size, the reduction in channel dimension adversely impacts detection performance. We conjecture that this is due to a decrease in the learning capability of each layer, leading to the propagation of misleading information to the multi-collaboration module, despite the presence of more pixel-level supervisions.
\begin{table}[h]
    \centering
    \caption{Ablation study on trade off between the spatial size and channel dimension.}
    \setlength{\tabcolsep}{0.25cm}
    \begin{tabular}{cc|cc|cc}
        \hline
        \multicolumn{2}{c|}{Combination} & \multicolumn{2}{c|}{Facial Components} & \multicolumn{2}{c}{Facial Attributes}\\
        \cline{1-6}
         spatial size & channel & Ada-ACC & CSM  & Ada-ACC & CSM\\
         \hline
         \hline
        256&64 &	52.10&46.26&49.05	&51.69	\\
        128&128 &	52.52&46.63&47.79	&50.65	\\
        64&256 &	\textbf{53.04}&	\textbf{47.19}&\textbf{49.30}&	\textbf{51.90}	\\
        \hline
    \end{tabular}
    \label{8}
\end{table}
\subsubsection{Discussion on the value setting in the manipulated mask}  In order to further investigate the appropriate value settings for the manipulated mask, Fig. \ref{f6}(a) provides insight into the values of the large manipulated region. A crucial consideration in the detection of manipulated regions is the selection of an appropriate value for the large manipulated region. Assigning lower values may diminish the significance of the large manipulated region. On the other hand, higher values can diminish the impact of small manipulated regions, which can negatively affect the overall detection performance. Striking the right balance in value assignment is imperative to ensure accurate and reliable detection outcomes.

\subsubsection{Discussion on the binary deepfake detection} 
\begin{figure}
\includegraphics[width=\linewidth]{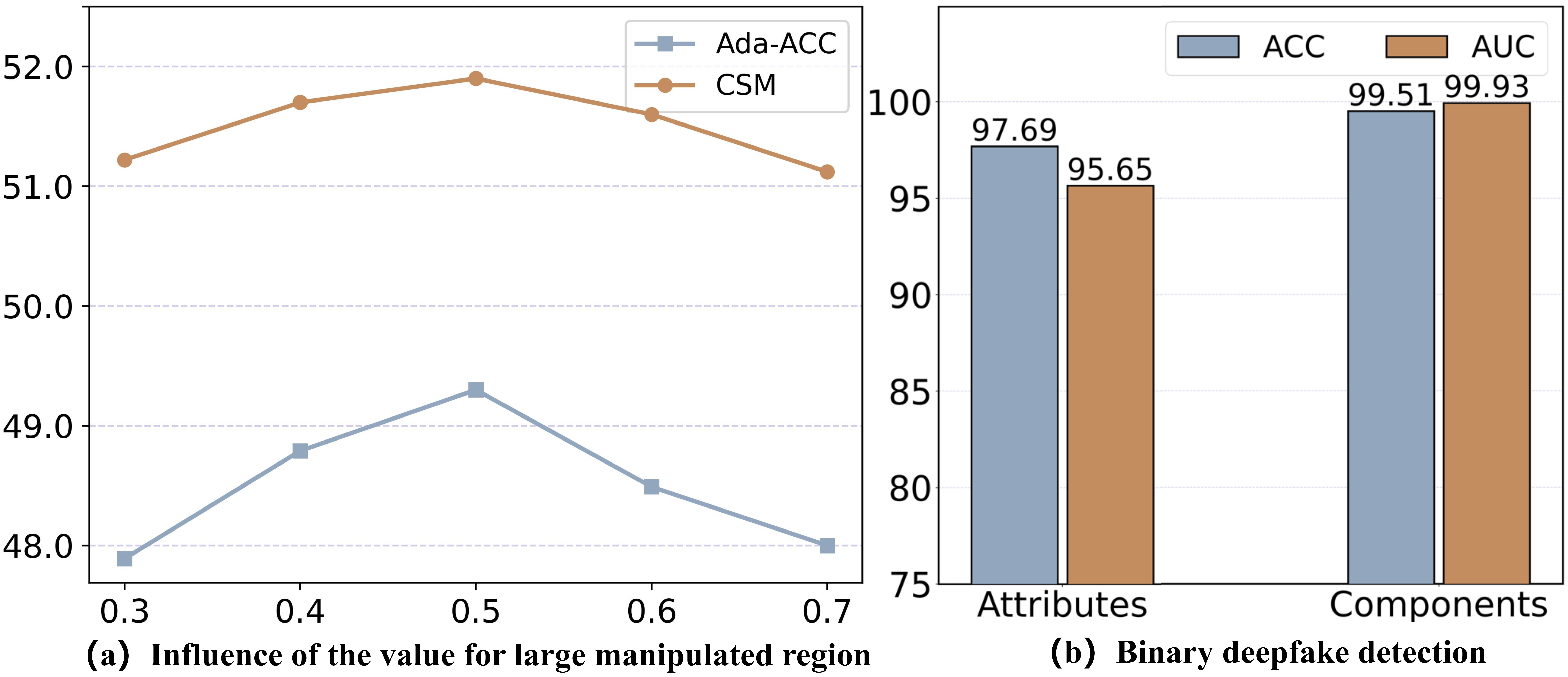}
\caption{(a) The influence of value setting for the large manipulated region on the facial attribute validation set. (b) The binary deepfake detection performance for MMNet.}
\label{f6}
\end{figure}

To delve into the identification of authenticity, we consider the face images without/with sequential manipulations as real/fake. The results in Fig. \ref{f6}(b) show that the proposed model can respectively achieve 95.65\% and 99.93\% AUC (Area Under Curve) on facial attributes and components test sets, which demonstrates the availability of our model on binary deepfake detection. The numerical difference between the binary accuracy and sequential adaptive accuracy indicates the hardness of sequential deepfake detection as the model needs to output the specific manipulated region labels with the correct sequence. Moreover, the high binary accuracy ensures the feasibility of the subsequent recovery framework as it only focuses on the forged face images.

\subsection{Evaluation on the recovery performance}
\subsubsection{Discussion on the independent recovery performance} To evaluate the recovery performance of the proposed recovery framework, the discrepancies between the fake and real face images are shown in the first row of Table \ref{4}. The aim of the recovery framework is to generate images that are more close to real ones. Four classic image similarity evaluation metrics are listed to make the comparison of both the pixel-level and feature perception-level differences.
\par
\begin{figure}
\centering
\includegraphics[width=\linewidth]{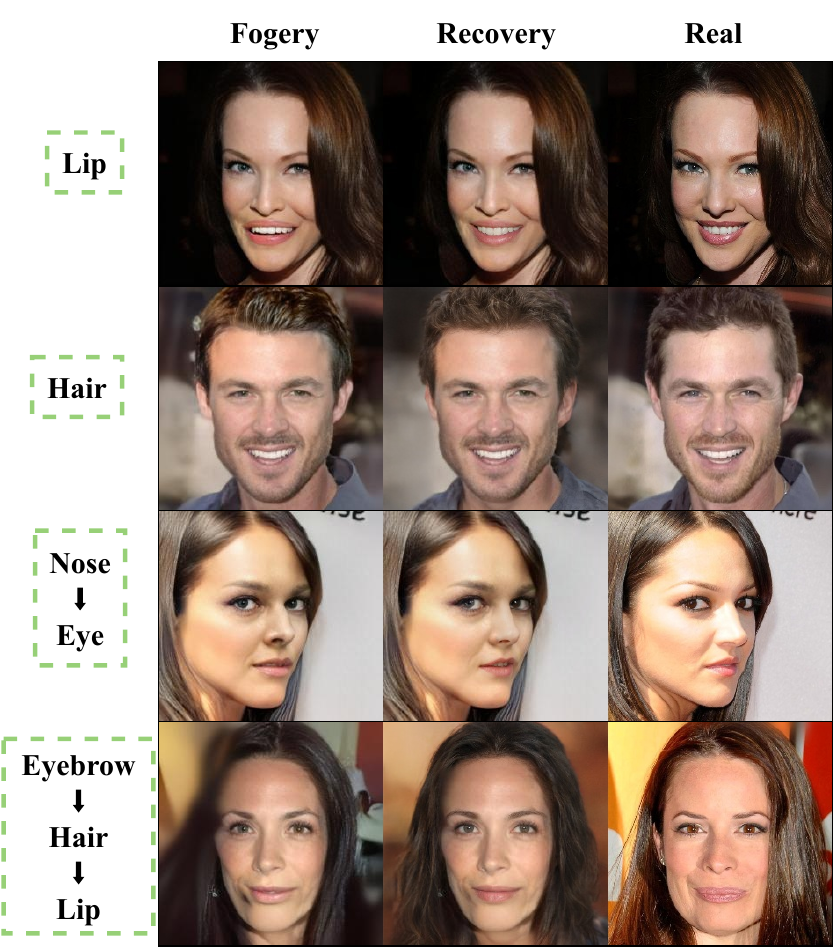}
\caption{Qualitative results are sampled from the facial component dataset. The contents from the left of the forgery are the sequential ground-truth.}
\label{f7}
\end{figure}

 Based on the quantitative results, it can be observed that face images generated from the multi-supervision module with SRM show a reduction in pixel-level differences to some extent in the second row of Table \ref{4}. However, a decrease in Structural Similarity Index (SSIM), Peak Signal-to-Noise Ratio (PSNR), and an increase in Learned Perceptual Image Patch Similarity (LPIPS) indicate that the overall perceptual differences, such as luminance, contrast, and structure, are enlarged. This is attributed to the forged face images exhibiting different manipulation traces, making them challenging for the model to directly recover. Hence, our proposed recovery framework takes into consideration the local regions of the images generated from SRM as the foreground, while treating the remaining regions in the forged face image as the background because the editing technique mainly focuses on the designated regions. In comparison to the forgeries, in the third row of Table \ref{4}, images generated from the proposed recovery framework demonstrate a significant improvement in both pixel and perceptual similarity. Fig. \ref{f7} illustrates the translation from forgery to real images, showcasing that the recoveries are much closer to the real images, thus validating the effectiveness of our proposed recovery framework.
\par It is a worthy note that our proposed framework focuses on recovering forged face images without the knowledge of the manipulation model, which sets it apart from the comparison with the recovery process when the model is known. Secondly, the comparison is impeded by the lack of corresponding real images in the facial attributes dataset, which limits the availability of relevant recovery results \cite{6}. Furthermore, the recovery results shown in \cite{6} only involve 100 randomly selected samples. In contrast, our work provides statistical results encompassing the entire test set, offering a more comprehensive and representative evaluation.

\begin{table}
    \centering
    \caption{Comparison of the difference from the real image to the generated images with different models.}
    \resizebox{\columnwidth}{!}{
    \begin{tabular}{c|cccc}
        \hline
        \multirow{2}{*}{Model}  & \multicolumn{4}{c}{Facial Components}\\
        \cline{2-5}
         & MSE $\downarrow$ & SSIM$\uparrow$ & PSNR$\uparrow$ & LPIPS$\downarrow$\\
         \hline
         \hline
         Manipulation model &0.029 & 0.535 & 17.57 & 0.264\\
         Multi-Supervision module with SRM &0.026 & 0.454 & 16.45 & 0.284\\
         Recovery framework &\textbf{0.022} & \textbf{0.537} & \textbf{18.11} & \textbf{0.249}\\
         \hline
         SRM (downsampled real image) &0.015 & 0.543 & 18.71& 0.234\\
         \hline
    \end{tabular}}
    \label{4}
\end{table}

\subsubsection{Discussion on the local recovery performance} To assess the efficacy of the recovery process on a local level, we respectively compare the local regions of the fake and recovered face images w.r.t the real ones. These analyses are presented in Table \ref{10}. Apparently, the large manipulated regions (i.e., hair) are more prone to deviate from the real images, and hence more difficult to recover than the small manipulated regions (i.e., nose, eye, eyebrow, and lip). Nonetheless, the proposed recovery framework can still make the recovered regions resemble the real ones. To comprehensively evaluate the local recovery performance, we average the results across all manipulated facial components. The final row of Table \ref{10} shows that our recovery framework can significantly elevate the similarity metrics, including MSE, SSIM, and PSNR, thereby indicating its efficacy in bridging the gap between the recovered and real images.
\begin{table}[h]
    \centering
    \caption{Comparison of the differences between real vs. fake and real vs. recovery.}
    \resizebox{\columnwidth}{!}{
    \begin{tabular}{c|ccc|ccc}
        \hline
        \multirow{2}{*}{Components}  & \multicolumn{3}{c}{Real vs. Fake} & \multicolumn{3}{|c}{Real vs. Recovery}\\
        \cline{2-7}
         & MSE $\downarrow$ & SSIM$\uparrow$ & PSNR$\uparrow$& MSE $\downarrow$ & SSIM$\uparrow$ & PSNR$\uparrow$\\
         \hline
         \hline
         Nose &	0.027	& 0.520	& 16.86 & 0.018	&0.603 & 18.49	\\
         Eye & 0.030 & 0.533 & 15.89 & 0.027& 0.541 & 16.22 \\
         Eyebrow &  0.029 & 0.488 & 16.20&0.026&  0.501&  16.72  \\
         Lip & 0.027 & 0.426 & 16.38&0.023 &	0.428 &	16.93\\
         Hair &	0.065 &	0.371 &	13.34&0.045	& 0.378& 14.72\\
         \hline
         Mean & 0.036 & 0.467 & 15.73 &0.027 & 0.490 & 16.61\\
         \hline
    \end{tabular}}
    \label{10}
\end{table}

\begin{figure*}[ht]
\includegraphics[width=\linewidth]{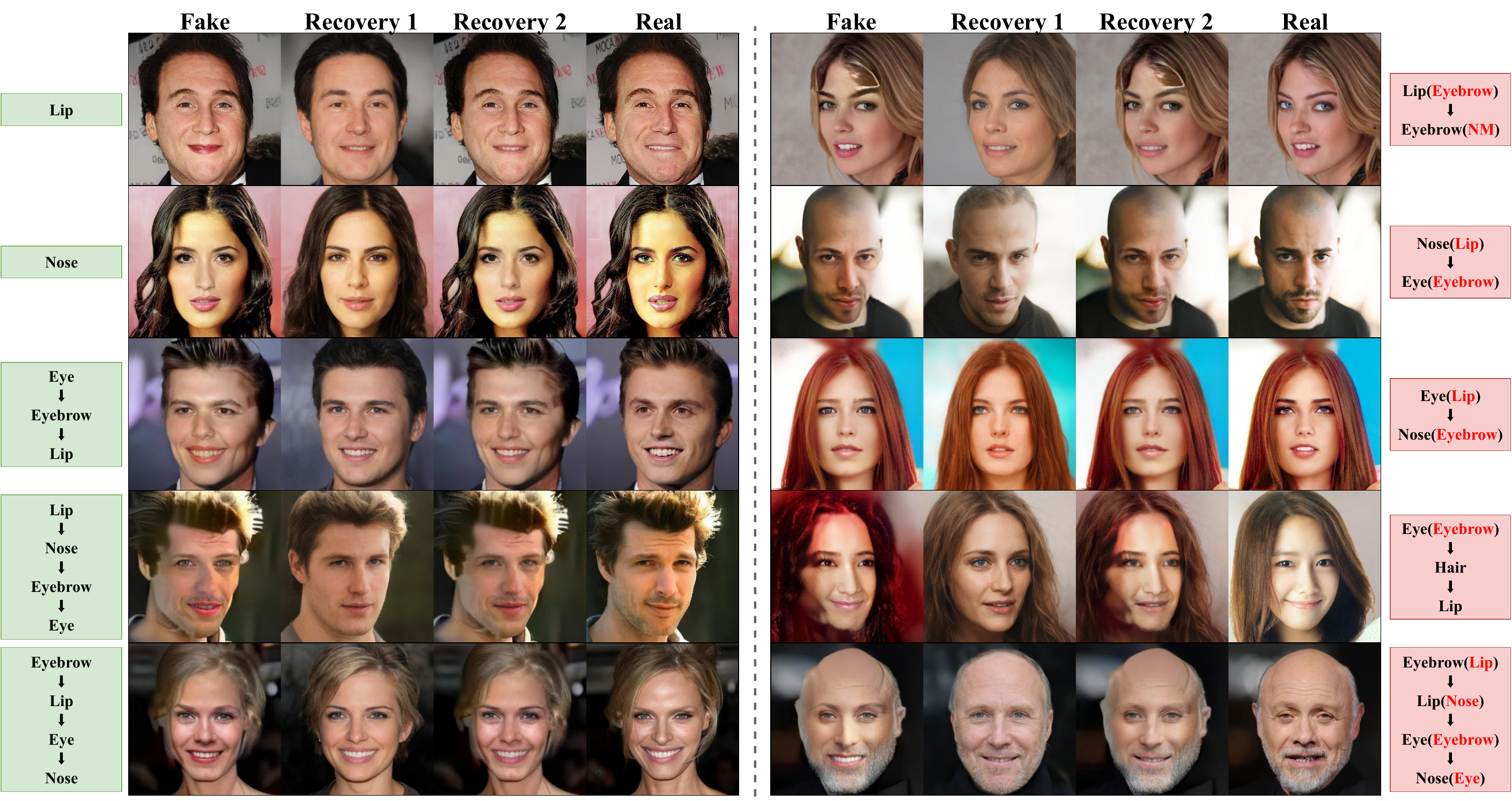}
\centering
\caption{More qualitative results. Recovery 1 and 2 denote the recovered images from the multi-supervision module with SRM and the proposed recovery framework, respectively. In the wrong predictions (red box), the manipulated region labels in red font indicate the ground-truth.}
\label{f11}
\end{figure*}
\subsubsection{Discussion on the recovery error} Although the recovery framework has shown promising results in recovering forged face images from their real counterparts, there are still some errors present in the recovered images. These errors can be attributed to three main factors: the SRM, the multi-supervision module, and the manipulated label prediction. 
\par An experiment is conducted where we employ a split approach for the SRM by inputting the real images with the same scale as the output of the multi-supervision module. The differences observed between the super-resolution images and the real images are listed in the last row of Table \ref{4}, which indicates the super-resolution error of SRM. This is likely due to the pretrained domain bias (trained from FFHQ) and the noise amplification caused during the clarification process. 
\par Furthermore, the metric differences between the second and last row in Table \ref{4} represent the errors between the images generated from the multi-supervision module and the downsampled real images. These errors stem from the limited learned representation of the multi-supervision module, which may not fully capture the diverse manipulation traces present in forged face images. 
\par Additionally, our findings in Table \ref{5} shed light on the impact of manipulated label predictions on recovery performance. Wrong (sequence) means correct predicted labels with the wrong sequence. Wrong (half) means at least half of the labels in the predicted sequence match with the sequential ground-truth. Correct means all-time predictions are the same as the sequential ground-truth. The experimental results indicate that accurate predictions play a vital role in improving overall recovery performance. Specifically, correct predictions contribute to enhanced recovery outcomes by ensuring unbiased blending masks. On the other hand, incorrect predictions in wrong (half) can lead to biased blending masks, which can have a detrimental impact on recovery performance. Moreover, as only the multi-supervision module pays attention to the sequential manipulation traces in our proposed recovery framework, the results from the wrong (sequence) indicate that even if the blending mask is correct, the variation of sequential manipulation traces can still make the recovery performance worse. Therefore, the investigations and refinement of these aspects will be the future work to enhance the recovery performance. More qualitative results are shown in Fig. \ref{f11}. 

\begin{table}
    \centering
    \caption{Comparison among the difference from the recoveries with correct and wrong predictions.}
    \resizebox{\columnwidth}{!}{
    \begin{tabular}{c|cccc}
        \hline
        \multirow{2}{*}{Prediction}  & \multicolumn{4}{c}{Facial Components}\\
        \cline{2-5}
         & MSE $\downarrow$ & SSIM$\uparrow$ & PSNR$\uparrow$ & LPIPS$\downarrow$\\
         \hline
         \hline
         wrong (sequence) &{0.027} & {0.474} & {16.69} & {0.284}\\
         wrong (half) &{0.026} & {0.507} & {17.08} & {0.278}\\
         correct &{\textbf{0.020}} &{\textbf{0.551}} & {\textbf{18.63}} & {\textbf{0.238}}\\
         \hline
         all &{0.022} & {0.537} & {18.11} & {0.249}\\
         \hline
    \end{tabular}}
    \label{5}
\end{table}

\section{Conclusion}
\par In this paper, we have introduced Multi-Collaboration and Multi-Supervision Network (MMNet) for sequential deepfake detection. The proposed detection framework of MMNet is designed to effectively handle various manipulation traces commonly found in forged images. The proposed recovery framework is the early exploration to achieve independent face recovery. Additionally, we have introduced a new evaluation metric called Complete Sequence Matching (CSM) to assess the detection accuracy of the model over continuous multiple steps. Our experimental results on several public datasets have shown the superiority of MMNet over existing methods, and the necessity of incorporating the proposed evaluation metrics. In the future, our work will major in improving the multi-supervision module, SRM, and sequential predictions, which are closely related to detection as well as recovery performance.



%





\ifCLASSOPTIONcaptionsoff
  \newpage
\fi





\bibliographystyle{IEEEtran}
\bibliography{manuscript}

\vfill


\end{document}